# *Training and Profiling a Pediatric Emotion Recognition Classifier on Mobile Devices*

Agnik Banerjee
N/A
Lynbrook High School
San Jose, California
agnik.banerjee@outlook.com

Peter Washington
Bioengineering
Stanford University
Stanford, California
peterwashington@stanford.edu

Cezmi Mutlu
Electrical Engineering
Stanford University
Stanford, California
cezmi@stanford.edu

Aaron Kline
Pediatrics (Systems Medicine)
Stanford University
Stanford, California
akline@stanford.edu

Dennis P. Wall*
Pediatrics (Systems Medicine), Biomedical Data Science
Stanford University
Stanford, California
dpwall@stanford.edu

*Abstract*—Implementing automated emotion recognition on mobile devices could provide an accessible diagnostic and therapeutic tool for those who struggle to recognize emotion, including children with developmental behavioral conditions such as autism. Although recent advances have been made in building more accurate emotion classifiers, existing models are too computationally expensive to be deployed on mobile devices. In this study, we optimized and profiled various machine learning models designed for inference on edge devices and were able to match previous state of the art results for emotion recognition on children. Our best model, a MobileNet-V2 network pre-trained on ImageNet, achieved 65.11% balanced accuracy and 64.19% F1-score on CAFE, while achieving a 45-millisecond inference latency on a Motorola Moto G6 phone. This balanced accuracy is only 1.79% less than the current state of the art for CAFE, which used a model that contains 26.62x more parameters and was unable to run on the Moto G6, even when fully optimized. This work validates that with specialized design and optimization techniques, machine learning models can become lightweight enough for deployment on mobile devices and still achieve high accuracies on difficult image classification tasks.

## I. INTRODUCTION

Autism Spectrum Disorder (ASD) affects 1 in 54 children and is the fastest growing developmental disability in the U.S., with its prevalence having increased by 178% since 2000 [1]. While research has shown that early detection and therapy is vital in order to treat ASD [2-3], a lack of access to clinical practitioners, particularly among lower-income families [4-5], results in 27% of children over the age of 8 remaining undiagnosed and too old to respond as effectively to treatment in the future [6].

Due to lowering prices, digital technologies are becoming widely available in almost all socioeconomic levels [7], even in developing nations where ASD clinicians are far scarcer than the U.S. [8]. Thus, it is conceivable that mobile apps on smartphones could be used as an alternative medium for autism diagnosis and treatment that is easily accessible and highly affordable.

Machine learning models utilized in digital therapeutics have shown significant diagnostic [9-25] and treatment [26-32] capabilities for children with ASD. These models rely on large volumes of data, however, and with children being severely underrepresented in the few datasets available, there is a necessity for more behavioral data for both neurotypical and autistic children [34]. To address this issue, we have previously developed a Charades-style mobile game named GuessWhat [35-39], which challenges children with ASD to improve their social interactions while simultaneously collecting structured data for diagnostic and therapeutic AI development.

Because children with ASD and similar developmental behavioral conditions often display significant impairment in both the understanding and imitation of facial emotion [68-69], we explored the deployment of emotion recognition classifiers on mobile devices in this study. We optimized for both classification performance and efficiency on a Motorola Moto G6 phone. Our best model was able to match previous state of the art results on emotion recognition for children while being lightweight enough to perform inference on the Moto G6 in real-time. This work can be used for a variety of applications such as mobile health therapies for ASD which provide targeted emotion treatment based on the affective profile of the user.

## II. RELATED WORK

The fields of facial emotion recognition and development of machine learning models for edge devices are vast. Prior work relevant to this study can be divided into three categories: (A) facial emotion recognition, (B) neural architecture search, and (C) model compression techniques.

## A. Facial Emotion Recognition

Facial emotion recognition (FER) is a widely researched field with a large library of datasets and classifiers. Early techniques introduced by Kharat et al. [40] involved extracting facial key points from faces and passing them through standard models such as support vector machines. Although initial results using this method were promising and computationally inexpensive, these classifiers were evaluated against small, well-structured datasets such as the Extended Cohn-Kanade (CK+) dataset [41] and the Japanese Female Facial Expression (JAFFE) dataset [42]. When tested against more heterogeneous data of images taken from a variety of orientations, such as the Face Expression Recognition 2013 (FER-2013) dataset [43], models received much lower scores [44].

Out of all existing techniques, convolutional neural networks (CNNs) have shown the greatest potential in both accuracy and generalizability due to their powerful automatic feature extraction. [45-46]. Thus, CNNs are presently the most widely used technique in FER, with an ensemble of CNNs with residual masking blocks leveraged by Luan et al. to achieve current state of the art results [47].

While CNNs have seen results in FER improve consistently throughout recent years and have been used in similar applications such as eye gaze detection [70-71], there are few endeavors involving classification on children's faces. The Child Affective Facial Expression Set (CAFE) [48] currently is the largest dataset of facial expressions from children and a standard benchmark in the field of FER on children. The current state of the art on this dataset, achieved by Washington et al., was able to achieve 69% accuracy using a ResNet152-V2 architecture pre-trained on ImageNet weights [34].

## B. Neural Architecture Search

Neural architecture search (NAS) is a paradigm for several techniques to automate network architecture engineering. NAS can be used to find efficient deep neural network architectures that can be used for facial emotion recognition. While NAS takes a large amount of computational power to find the optimal network, it can be easily tailored to find the best model for a specific use case. For instance, Lee et al. used NAS to build EmotionNet Nano [49], which was able to outperform other state-of-the-art models at FER while optimizing for speed and energy. Although we were unable to pursue NAS in this study due to computational limitations, we highlight this field as an interesting area of potential research, especially when paired with model compression techniques.

## C. Model Compression Techniques

With CNNs being both computationally and memory intensive, several model compression techniques have been developed to make these models more lightweight post-training. Han et al. proposes three techniques to increase inference speed while decreasing memory overhead and energy consumption: weight pruning, weight clustering, and quantization [50].

Weight pruning involves gradually zeroing the magnitudes of the weights, making the model sparser by effectively removing weights that have the least significance in the model's predictions. When weight pruning is used with weight clustering, which groups homogeneous weights together to share common values, model size can be decreased by as much as 9x to 13x with negligible accuracy loss [50]. By quantizing the standard 32-bit weights of a model to a lower bit representation, models can be further compressed and even deployed on specialized edge hardware for faster inference [51]. In this study, we use all these techniques in conjunction to improve our models' performance.

## III. METHODS

### A. Data Collection

We leveraged images from ten relatively small yet well controlled datasets in order to train our models: NIMH Child Emotional Faces Picture Set (NIMH‑ChEFS) [53], Facial Expression Phoenix (FePh) [52], Karolinska Directed Emotional Faces (KDEF) [54], Averaged KDEF (AKDEF) [54], Dartmouth Database of Children's Faces (Dartmouth) [55], Extended Cohn-Kanade Dataset (CK+) [41], Japanese Female Facial Expression (JAFFE) [42], Radboud Faces Dataset (RaFD) [56], NimStim Set of Facial Expressions (NimStim) [57], and the Tsinghua Facial Expression Database (Tsinghua-FED) [58].

We also used the Face Expression Recognition 2013 (FER-2013) dataset [44] and a ~1600 subset of images from Expression in-the-Wild (ExpW) [59], a large library of web scraped images of faces, to balance the ratio of samples of each emotion. In total, 57884 images were used to train each model. Although we utilized a wider variety of datasets than the current state of the art, we used ~15000 fewer images [34].

While detailed background information was not provided for the subjects present in FePH, FER-2013 and ExpW, we were able to compile demographics for the remaining datasets, which are shown in the table below.

TABLE I. TRAINING DATA DEMOGRAPHICS

| Dataset | # Subjects | Age | Ethnicity | % Female |
|---|---|---|---|---|
| NIMH-ChEFS | 59 | M=13.57 SD=1.66 | Mostly Caucasian | 66.1 |
| KDEF/AKDEF | 70 | M=23.73 SD=7.24 | Latino | 50.0 |
| Dartmouth | 80 | M=9.84 SD=2.33 | Caucasian | 50.0 |
| CK+ | 123 | 98.14 | 81% Caucasian; 13% African | 69.00 |

| | | | American; 6% Other | |
|---|---|---|---|---|
| JAFFE | 10 | N/A | Asian | 100 |
| RaFD | 49 | M=21.2 SD=4.0 | Caucasian | 51.02 |
| NimStim | 43 | M=19.4 SD=1.2 | 58% Caucasian; 23% Afro-American; 14% Asian; 5% Latino | 41.86 |
| Tsinghua-FED | Group A: 67 Group B: 70 | Group A: M=23.82 SD=4.18 Group B: M=64.40 SD=3.51 | Asian | Group A: 50.75% Group B: 50% |

Despite collecting a large and diverse collection of datasets for our study, children were still heavily underrepresented in our training data with only NIMH-ChEFS, Dartmouth, and a small portion of RaFD containing children's faces. Although subjects came from a wide background of ethnicities, there still was an overwhelmingly large number of Caucasians and little to no presence of subjects from African or South-Asian descent. In the future, we hope that subsequent datasets manage to capture these underrepresented groups.

### B. Data Preprocessing

Before training our models, faces were cropped from all images using the Oxford VGGFace model [60] with a ResNet50 backbone. Images were then resized to 224x224 pixels and grayscale images were converted to three color channels. All images were then normalized to a range from -1 to 1.

### C. Model Training

We trained and compared five existing CNN architectures designed for deployment on mobile devices: MobileNetV3-Small 1.0x [61], MobileNetV2 1.0x [62], EfficientNet-B0 [63], MobileNetV3 1.0x [62], and NasNetMobile [64], all of which were pre-trained on ImageNet [65]. We retrained each layer of each network using categorical cross entropy loss and an Adam optimizer [66] with a learning rate of 1e-5. During training, all images were subject to a potential horizontal flip, zoomed in/out by a factor up to 0.15, rotated between -45 degrees and 45 degrees, shifted by a factor up to 0.10, and brightened by a factor between 0.80 and 1.20. We assumed the model converged and thus interrupted training once the validation loss did not improve for five consecutive epochs.

### D. Model Evaluation

We evaluated our models against the Child Affective Facial Expression Set (CAFE) [67], a large dataset consisting of facial expressions for children. CAFE is an excellent benchmark for this study as the children present are from a wide range of ethnic backgrounds, as shown in the figure below. CAFE's subjects are also aged between 2 to 8 years, the same range where an autism diagnosis is most vital [6]. Children in this dataset express seven emotions: happiness, sadness, surprise, fear, anger, disgust, and neutrality.

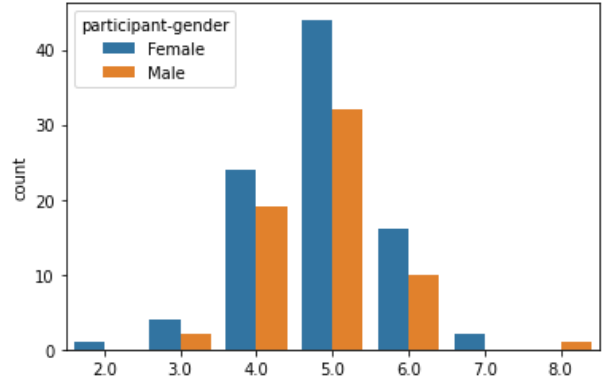

Fig. 1.    Gender and Age of CAFE Subjects

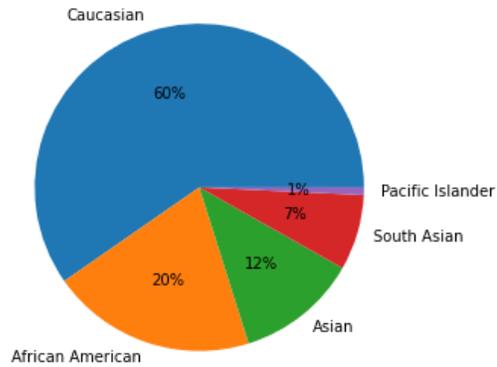

Fig. 2.    Ethnicity of CAFE Subjects

We additionally evaluated Subset A and Subset B of CAFE to observe our models' performance against faces that even human annotators have difficulty classifying. Subset A contains faces that were identified with 60% accuracy or above by 100 adult participants. In contrast, Subset B contains faces with substantially greater variability for each emotion, resulting in a Cronbach's alpha internal consistency score that is 0.052 lower than Subset A. [48].

We profiled all models on a Motorola Moto G6 Phone using the Tensorflow Lite benchmark API. We also deployed our models on an Android demo app we built that performs real-time image classification on a live video feed to ensure our models matched the results we received from the benchmark tool.

*E. Model Optimization and Reevaluation*

After evaluating on CAFE, we performed weight pruning before fine tuning the network until the validation loss did not improve for five consecutive epochs. We then applied weight clustering before fine tuning the network again in an identical fashion. We finally performed quantized-aware training before evaluating the fully optimized model against CAFE. If the model was unable to undergo quantized-aware training, we applied post-training quantization instead.

## IV. RESULTS

Upon evaluation, our best model was the MobileNetV2 1.0x, which achieved 65.11% balanced accuracy and 64.19% F1-score on CAFE (confusion matrix in Figure 3). This performance increases to 79.54% balanced accuracy and 79.43% F1-score on Subset A of CAFE (confusion matrix in Figure 4). When evaluated on CAFE Subset B, the MobileNetV2 model achieves 63.94% balanced accuracy and 62.32% F1-score (confusion matrix on Figure 5), attaining accuracies higher than those that even human annotators could achieve [48].

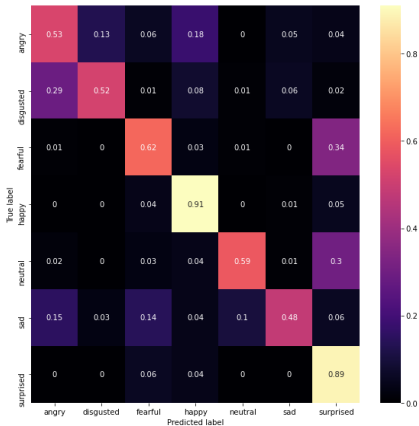

*Fig. 3.   Confusion Matrix for Entirety of CAFE*

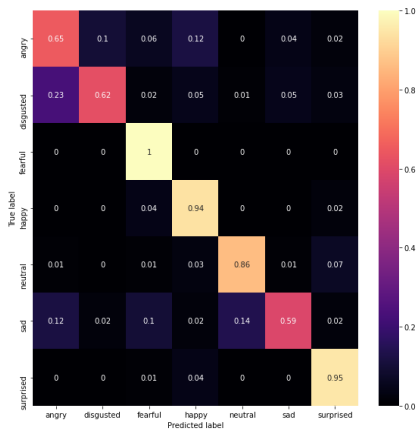

*Fig. 4.   Confusion Matrix for Subset A of CAFE*

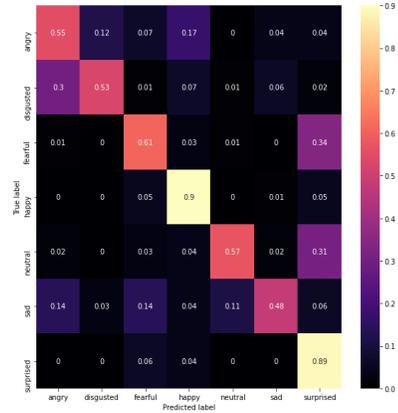

*Fig. 5.   Confusion Matrix for Subset B of CAFE*

Although models with more complex architectures than the MobileNetV2 were used, they performed slightly worse when evaluated on CAFE despite having higher validation accuracies. We believe that this issue occurs because the validation dataset still contains dissimilarities between the training dataset, even though we used the most diverse library of facial emotion images ever to train on CAFE. Thus, models with significantly more complexity overfit on the validation set and do worse when tested. Despite the discrepancy between the validation and testing set, all models still achieved impressive results with accuracies and F1-scores above 61%, nearly matching state of the art results while being far more lightweight.

TABLE I.  MODEL RESULTS

| Model | Size | | Balanced Accuracy (%) | | |
|---|---|---|---|---|---|
| | *Params (M)* | *FLOPs (M)* | *Subset A* | *Subset B* | *Total* |
| MobileNetV3-Small 1.0x | 1.54 | 85.1 | 75.11 | 62.38 | 63.13 |
| MobileNetV2 1.0x | 2.27 | 300 | 79.54 | 63.94 | 65.11 |
| EfficientNetB0 | 4.06 | 390 | 78.49 | 63.02 | 64.13 |
| MobileNetV3-Large 1.0x | 4.24 | 277 | 76.97 | 62.27 | 61.24 |
| NASNetMobile | 4.27 | 567 | 76.32 | 61.89 | 60.86 |

We profiled all five models on our Motorola Moto G6 phone and measured the memory consumption and latency when it performed inference on an image. We were able to decrease memory consumption by 4x and latency by ~1.3x using weight pruning, weight clustering, and quantization. These improvements are significant considering how few refinements could be made to these specific networks, as they already started out incredibly well-optimized simply by their architecture.

TABLE II. MODEL PERFORMANCE RESULTS

| Model | Original | | Optimized | |
|---|---|---|---|---|
| | Latency (ms) | Memory (mb) | Latency (ms) | Memory (mb) |
| MobileNetV3-Small 1.0x | 52.33 | 9.72 | 45.48 | 2.77 |
| MobileNetV2 1.0x | 62.33 | 13.78 | 45.61 | 4.11 |
| EfficientNetB0 | 415.04 | 19.91 | 301.62 | 6.08 |
| MobileNetV3-Large 1.0x | 124.47 | 14.82 | 98.14 | 4.34 |
| NASNetMobile | 218.07 | 26.98 | 192.85 | 8.99 |

a. Latency was recorded using 7 CPU threads

We also experimented with the optimal number of threads that should be added to the app process to get the best performance, and found that using seven threads on the Moto G6 yielded the lowest model latency for most models.

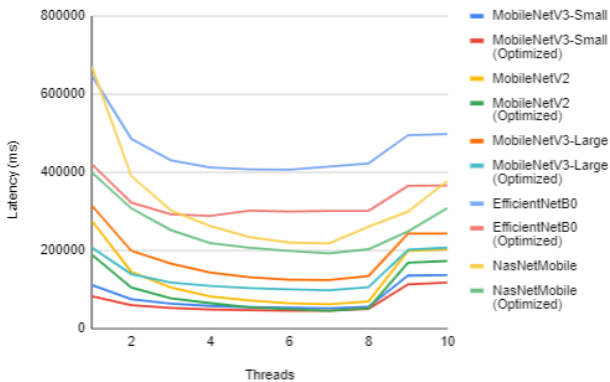

Fig. 6. Relationship between Model Latency and CPU Threads

## V. DISCUSSION

In this study, we trained several machine learning models to recognize emotion on childrens' faces. Using various optimization techniques, we were able to match state of the art accuracy while ensuring each model is able to perform real-time inference on a mobile device. These models are thus lightweight and accurate enough to be deployed in mobile health therapies such as targeted emotion treatment to children with ASD based on their affective profile, although specificity will need to be assessed in future studies. We also show that with specialized training, machine learning models designed for running on edge devices can achieve state of the art results on difficult classification tasks.

Although these results are promising, there were several limitations to this study. Due to a lack of computational resources, we were unable to create a custom model using neural architecture search. As a result, our model's backbone was not tailored specifically for the task of emotion classification on children and could still contain redundant layers despite being well optimized. In addition, while our models evaluated well on seven emotions, larger models may be needed to generalize to data with multiple and/or more emotions.

Future work includes the further collection of childrens' faces, which are still heavily underrepresented in facial emotion datasets and caused imbalance between the validation and testing datasets in this study. Another area of promise is deploying these models for on-device training. Once achieved, copies of a model can be sent to multiple decentralized mobile devices, before being concatenated as a global model once each model is individually trained on data captured by its device. This technique, known as federated learning, is an active field of research and could be leveraged to further improve these models in a privacy-preserving manner while they simultaneously provide therapeutic diagnosis and treatment of ASD.

ACKNOWLEDGEMENTS

This work was supported in part by funds to DPW from the National Institutes of Health (1R01EB025025-01, 1R21HD091500-01, 1R01LM013083, 1R01LM013364), the National Science Foundation (Award 2014232), The Hartwell Foundation, Bill and Melinda Gates Foundation, Coulter Foundation, Lucile Packard Foundation, the Weston Havens Foundation, and program grants from Stanford's Human Centered Artificial Intelligence Program, Stanford's Precision Health and Integrated Diagnostics Center (PHIND), Stanford's Beckman Center, Stanford's Bio-X Center, Predictives and Diagnostics Accelerator (SPADA) Spectrum, Stanford's Spark Program in Translational Research, Stanford mediaX, and Stanford's Wu Tsai Neurosciences Institute's Neuroscience: Translate Program. We also acknowledge generous support from David Orr, Imma Calvo, Bobby Dekesyer and Peter Sullivan. P.W. would like to acknowledge support from Mr. Schroeder and the Stanford Interdisciplinary Graduate Fellowship (SIGF) as the Schroeder Family Goldman Sachs Graduate Fellow.